\documentclass{article}
\usepackage{wrapfig}

\PassOptionsToPackage{numbers, compress, sort}{natbib}

\usepackage [ table ]{ xcolor }

\usepackage[final]{neurips_2023}

\usepackage[utf8]{inputenc} 
\usepackage[T1]{fontenc}    
\usepackage{hyperref}       
\usepackage{url}            
\usepackage{booktabs}       
\usepackage{amsfonts}       
\usepackage{nicefrac}       
\usepackage{microtype}      
\usepackage[colorinlistoftodos,bordercolor=orange,backgroundcolor=orange!20,linecolor=orange,textsize=scriptsize]{todonotes}

\usepackage{times}
\usepackage{mathtools}
\usepackage{latexsym}
\usepackage{algorithm}
\usepackage[noend]{algpseudocode}
\usepackage{mathrsfs}
\usepackage[T1]{fontenc}

\usepackage[utf8]{inputenc}
\usepackage{microtype}
\usepackage{amsthm}
\usepackage{graphicx}
\graphicspath{ {./images/} }
\usepackage{times}
\usepackage{latexsym}
\usepackage{amssymb}
\usepackage{amsmath}
\usepackage{bbm}
\usepackage{multirow}
\usepackage{inconsolata}
\usepackage{setspace}

\usepackage{xcolor, soul}

\newtheorem{theorem}{Theorem}[section]
\newtheorem{definition}[theorem]{Definition}
\title{READ: Recurrent Adaptation of Large Transformers}

\author{
  John Nguyen\thanks{Equal contribution. Correspondence emails: 
  \href{mailto:yuwang2020@meta.com}{yuwang2020@meta.com},
  \href{mailto:ngjhn@meta.com}{ngjhn@meta.com}
  } ~~~~~
  Sid Wang\footnotemark[1] ~~~~~
  Ke Li ~~~~~
  Carole-Jean Wu \\
  ~\\
  Meta\\
}

\begin{document}

\maketitle

\begin{abstract}

In the realm of Natural Language Processing (NLP), large-scale transformers have established themselves as pivotal, achieving unparalleled results across numerous tasks. The conventional approach involves pre-training these models on extensive web-scale data, followed by fine-tuning them for specific downstream tasks. However, the burgeoning size of these models, which has surged almost two orders of magnitude faster than GPU memory since 2018, has rendered their fine-tuning financially and computationally exorbitant, limiting this capability to a select few well-funded institutions. Parameter-efficient transfer learning (PETL) has emerged as a potential solution, aiming to efficiently adapt pre-trained model parameters to target tasks using smaller, task-specific models. Nonetheless, existing PETL methods either introduce additional inference latency or marginally reduce memory requirements during training, thus not fully addressing the primary motivation behind PETL. This paper introduces REcurrent ADaption (READ), a novel, lightweight, and memory-efficient fine-tuning method that incorporates a small RNN network alongside the backbone model. READ not only achieves comparable model quality to traditional fine-tuning, saving over 84\% in energy consumption, but also demonstrates scalability and independence from the backbone model size. Through extensive experiments on various NLP benchmarks, including the GLUE benchmark, READ showcases robust performance and high efficiency, reducing model training memory consumption by 56\% and GPU energy usage by 84\% relative to full-tuning, without significantly impacting inference latency and memory. We provide a theoretically justified, scalable solution for fine-tuning large transformers.

\end{abstract}

\section{Introduction}

\begin{figure}[hbt!]
    \centering
    \begin{minipage}[t]{0.32\linewidth}
        \centering
        \includegraphics[width=\linewidth]{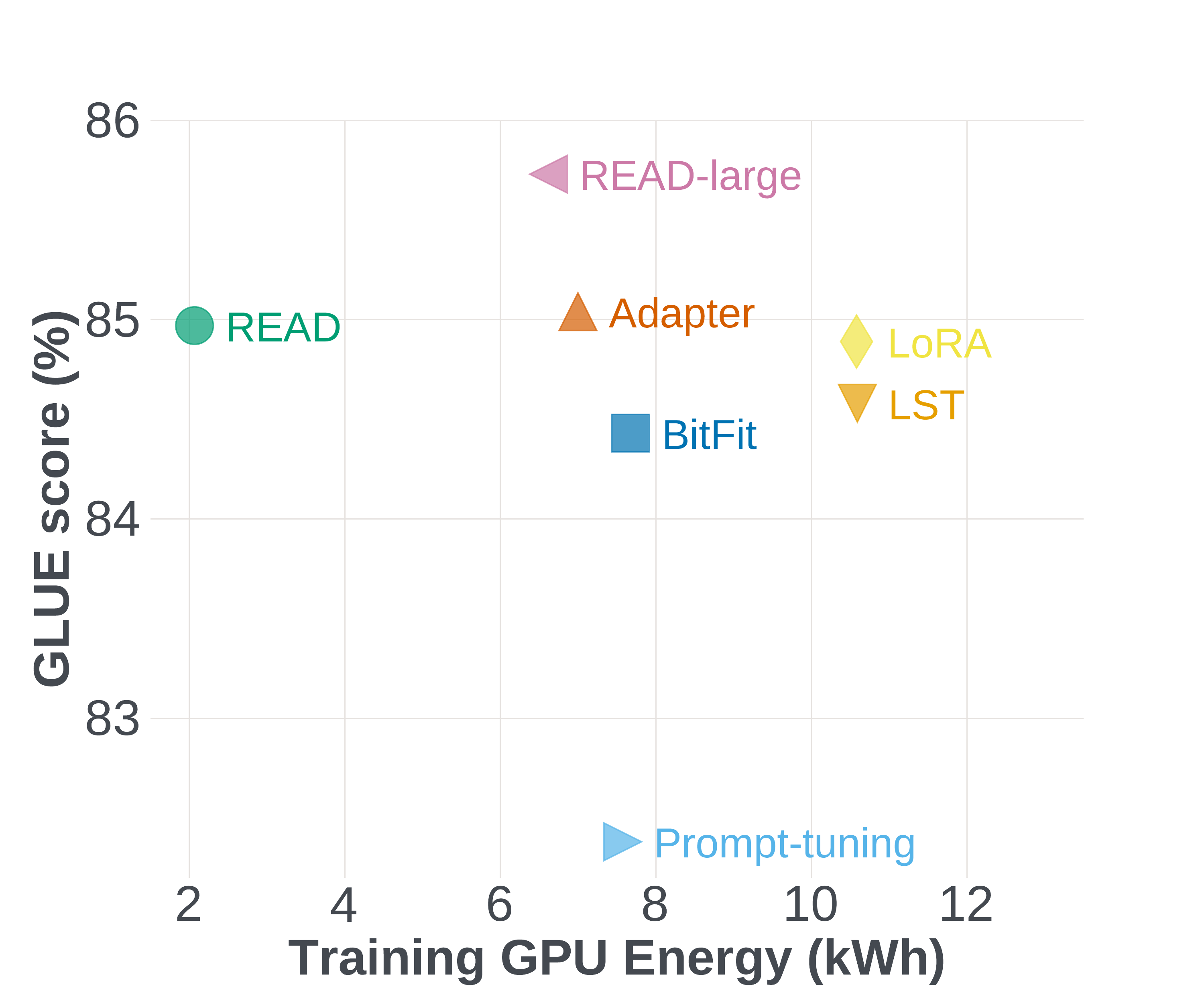}
    \end{minipage}
    \begin{minipage}[t]{0.32\linewidth}
        \centering
        \includegraphics[width=\linewidth]{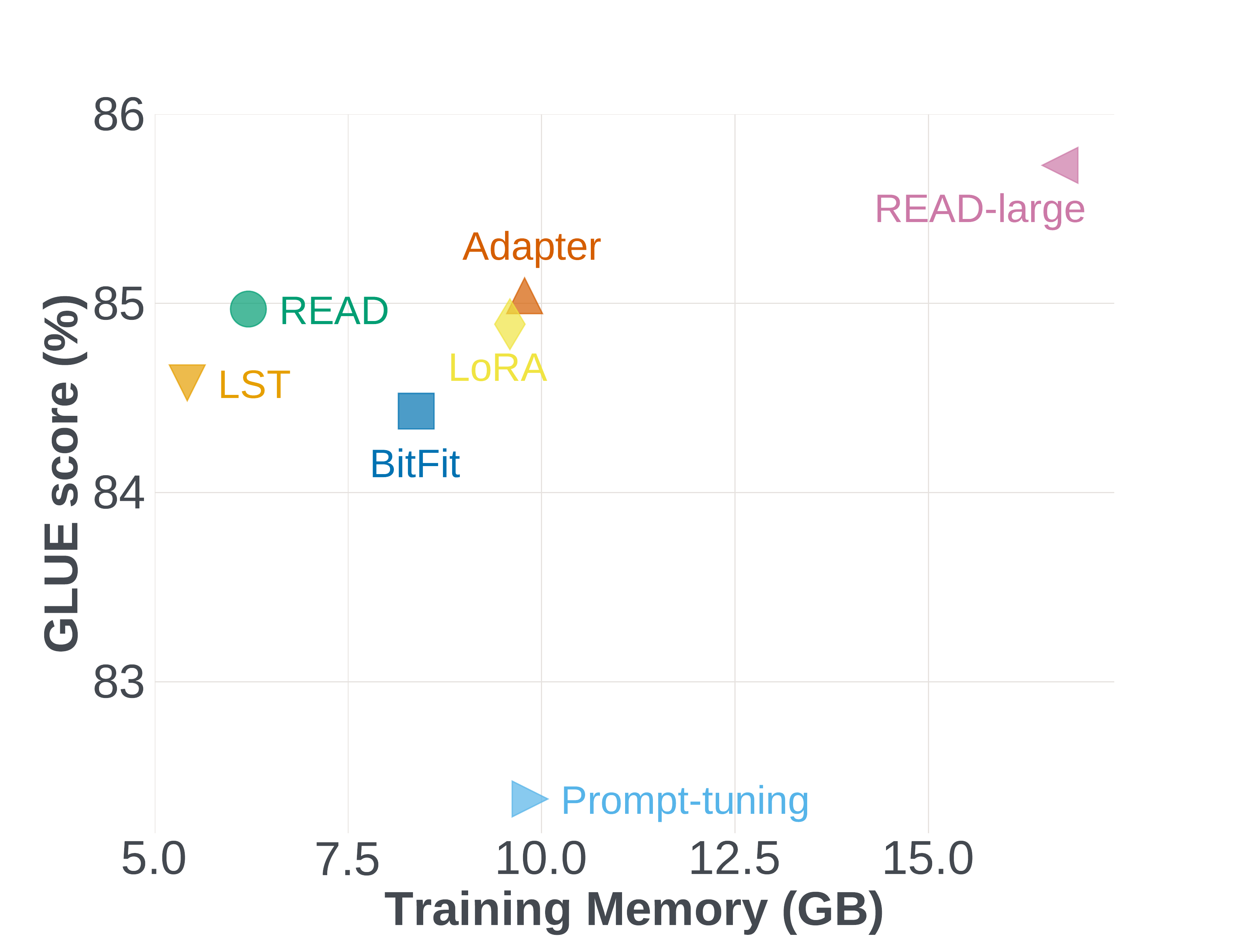}
    \end{minipage}
    \begin{minipage}[t]{0.32\linewidth}
        \centering
        \includegraphics[width=\linewidth]{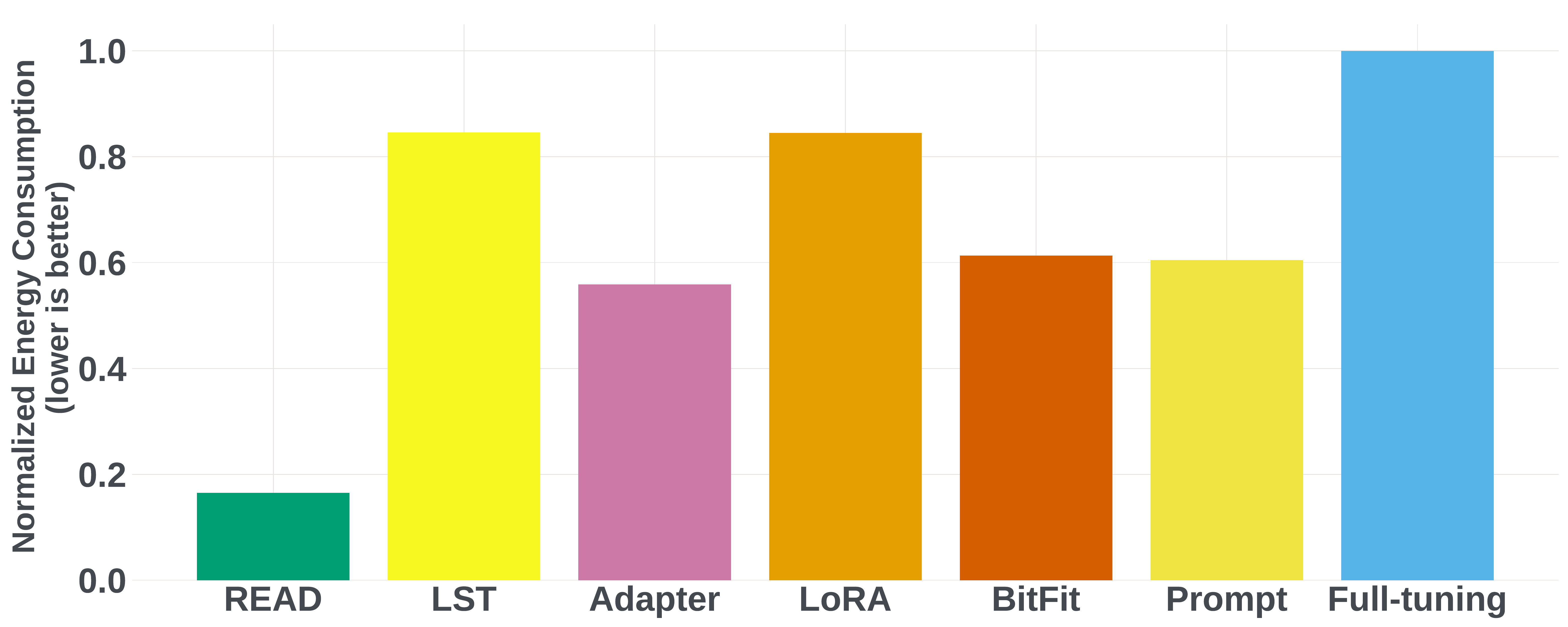}
    \end{minipage}
    \caption{(Left) Comparison of READ and other fine-tuning methods over GLUE tasks on training energy. (Center) Peak training memory relative to full-tuning. (Right) Normalized energy consumption relative to full-tuning on GLUE tasks.}
    \label{fig:combined_figures}
\end{figure}
Large-scale transformers architecture have achieved state-of-the-art results in several Natural Language Processing (NLP) tasks \cite{gpt_3, flan, T5, roberta, lu2019vilbert, bao2021beit}. Scaling up the size of these models has been shown to confer various benefits, such as improved model prediction performance and sample efficiency \cite{chung2022scaling, wei2023larger, howard2018universal}. The conventional paradigm is to pre-train large-scale models on generic web-scale data and fine-tune the models to downstream tasks. However, fine-tuning these models has become prohibitively expensive. 

Since 2018, the model size has increased by almost two orders of magnitude faster than GPU memory \cite{petl_survey}, resulting in prohibitively high cost to advance AI technologies~\cite{sustainable-ai}. Only a few well-funded institutions have the resources to fine-tune these models. Parameter-efficient transfer learning (PETL) \cite{adapters, lora, prefix, bitfit, prompt, compacter, aghajanyan2020intrinsic} has emerged as a promising solution to overcome the challenges of full fine-tuning. Parameter-efficient transfer learning techniques aim to address these challenges by leveraging smaller and more task-specific models to efficiently adapt the pre-trained model's parameters to the target task. 

However, all these methods either come with additional inference latency \cite{adapters} or reduces only a small amount of memory requirement during training --- the primary motivation of PETL. Figure \ref{fig:combined_figures} illustrates that parameter-efficient methods, while tuning only a small percentage of the overall parameters, still consume significant energy to fine-fine. Since the updated parameters are inside the backbone language models, to calculate gradients for these parameters for backpropagation, PETL methods need to run the backward pass through the sizeable pre-trained language models. This prevents PETL methods from being applied to many real-world applications with limited computational resources.
Recent works of Side-Tuning \cite{side_tuning} and Ladder-Side Tuning (LST) \cite{lst} propose to use a side network that takes intermediate activations from the backbone networks to reduce the need to backpropagate through the large backbone layer. It thus reduces training memory requirement. However, both Side-Tuning and LST have significant drawbacks. In Side-Tuning, the side network only consumes the original inputs, leaving the informative intermediate results from the backbone unused. LST overcomes this problem by using a side Transformer. However, Transformers are challenging to train \cite{Liu2020UnderstandingTD}. Moreover, LST requires an extra pretraining stage to extract a sub-Transformer from the backbone and use it to initialize the side network, increasing the cost of fine-tuning. Additionally, the size of the side-Transformer and the backbone increase, making this approach hard to scale (Figure \ref{adapted_size_backbone_size}). 
In this work, we introduce \textbf{RE}current \textbf{AD}aptation (READ), a novel, lightweight, and efficient fine-tuning method that incorporates a small Recurrent Neural Network (RNN) alongside the backbone model. READ not only achieves comparable model quality to traditional fine-tuning but also realizes more than 84\% energy savings during the process.

\textbf{Contributions}

We present READ, a memory and parameter-efficient fine-tuning method that:
\begin{enumerate}
    \item Introduces a side-tuning design that requires no pretraining of the side network, addressing limitations of previous PETL and side-tuning methods.
    \item Demonstrates robust performance and efficiency across various NLP benchmarks, reducing model training memory consumption by $56\%$ and GPU energy usage by $84\%$ relative to full-tuning.
    \item Proves scalable for fine-tuning large transformers, independent of the backbone model size.
    \item Offers theoretical justification for utilizing the backbone hidden state for side-tuning.
\end{enumerate}

\section{Breaking Down REcurrent ADaptation (READ)}
\subsection{What is READ?}
\label{sec:read}

\begin{figure}[htbp]
    \begin{minipage}[t]{0.49\linewidth}
        \centering
        \includegraphics[width=\linewidth]{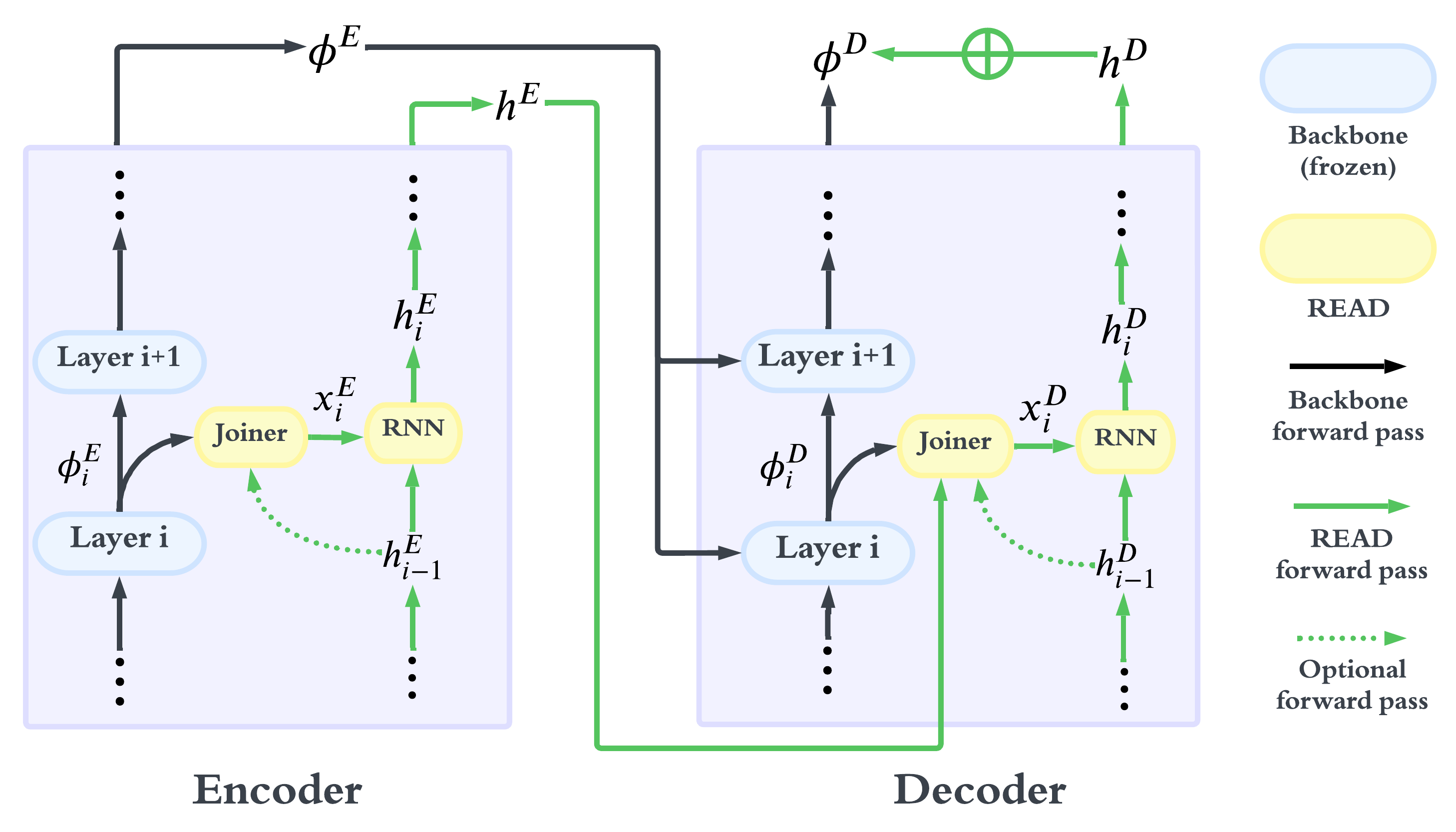}
        \caption{READ Fine-Tuning Mechanism.}
        \label{fig:read_arch}
    \end{minipage}
    \hfill 
    \begin{minipage}[t]{0.49\linewidth}
        \centering
        \includegraphics[width=\linewidth]{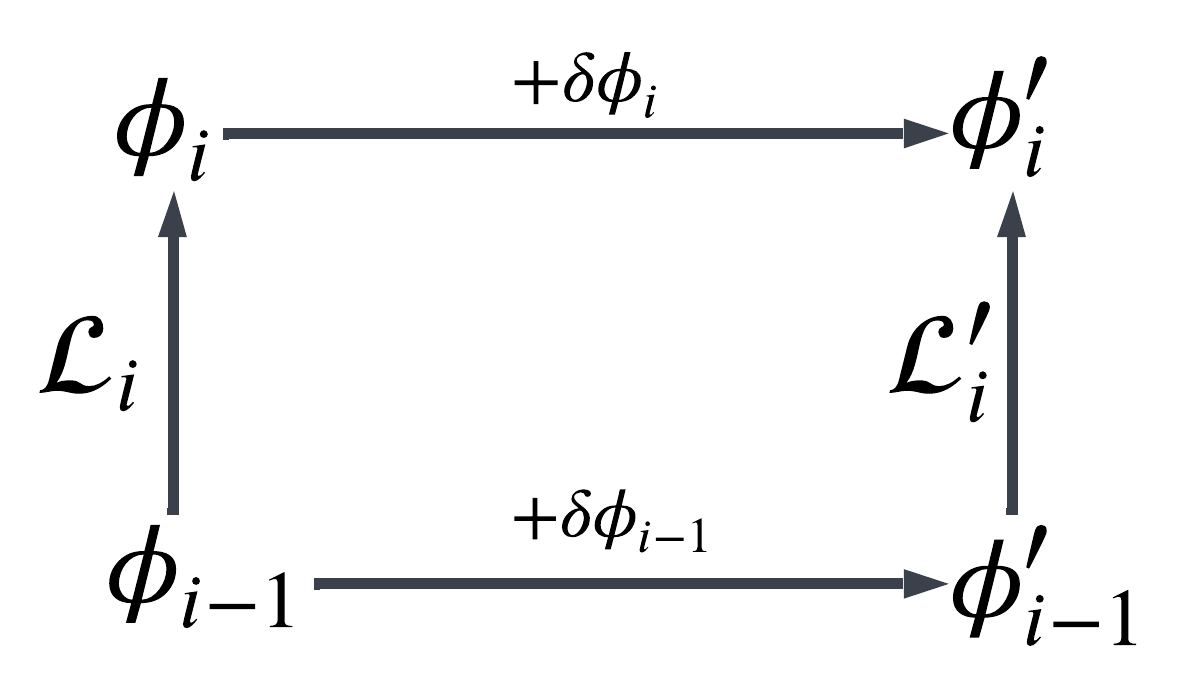}
        \caption{Commuting diagram for \emph{correction}.}
        \label{fig:comm}
    \end{minipage}
\end{figure}

Figure \ref{fig:read_arch} illustrates the READ fine-tuning mechanism on an encoder-decoder transformer backbone, \(\mathcal{T}\), which is frozen during training. READ, initialized at both encoder and decoder, primarily consists of a standard RNN and a \emph{Joiner} network, facilitating the amalgamation of multiple information sources to generate RNN inputs. The forward pass through \(\mathcal{T}\) is independent of READ, with intermediate results cached at each transformer layer, followed by iterative computation of RNN hidden states and the addition of RNN and \(\mathcal{T}\) outputs to derive the final state.

Key properties of READ include:
\begin{enumerate}
    \item \textbf{Isolation from Backbone}: Forward pass is separate from \(\mathcal{T}\), preventing backward propagation through it and reducing training memory \cite{lst}.
    \item \textbf{Simplicity and Efficiency}: Involves only RNNs and FFNs, enhancing usability and training efficiency without requiring pre-training.
    \item \textbf{Parameter Scalability}: The recurrent nature ensures trainable parameters do not increase with backbone layers, exhibiting sub-linear growth with backbone size.
    \item \textbf{Unmodified Intermediate Results Consumption}: READ utilizes without altering the backbone's intermediate results\footnote{Notably, although not detailed in this paper, READ enables \emph{multi-tasking} with multiple networks, necessitating only a single backbone pass, thereby reducing training and inference costs.}.
\end{enumerate}

\subsection{How does READ work?}
Figure \ref{fig:read_arch} provides a visual representation of the READ fine-tuning mechanism, applied to an encoder-decoder transformer backbone, denoted as \(\mathcal{T}\). In simpler terms, READ learns the necessary adjustments, or corrections, to the output hidden states at each layer of \(\mathcal{T}\) to adapt it for a new task.

\begin{definition}[\emph{Correction}]
\label{correction}
\emph{If \(\mathcal{T}^\prime\) is a modified version of \(\mathcal{T}\) and \(\phi_i^\prime\) represents the hidden states at layer \(\mathcal{L}^\prime_i\) of \(\mathcal{T}^\prime\), the difference \(\phi_i^\prime - \phi_i\) is termed a \emph{correction} to \(\phi_i\), and is denoted as \(\delta\phi_i\).}
\end{definition}

In essence, a correction (\(\delta\phi_i\)) represents the difference between the hidden states of the original and modified transformer layers, as illustrated in Figure \ref{fig:comm}.
\textbf{Key Insights into READ}

\textbf{Separation from Other Methods:} Unlike many fine-tuning methods that directly alter \(\phi_i\) by updating backbone weights or injecting learnable parameters, READ focuses on learning the \emph{correction} needed for a new task.

\textbf{Practical Application:} In practice, we utilize a neural network, named READ, to model the equation system. This involves employing a \emph{Joiner} network to compute \(x_i\), substituting various mathematical entities in the equation with Feed-Forward Networks (FFNs) or linear layers and merging learnable parameters across all indices \(i\) for efficiency and scalability. Additionally, Recurrent Neural Networks (RNN) are utilized to model part of the equation system.

This approach does not involve attention mechanisms and operates only on the column space of \(\phi\), ensuring that all operations are executed efficiently and effectively.

\section{Experiment Setup}
\textbf{Baseline and Other State-of-the-Art Designs}
We compare READ against full tuning and other commonly-used PETL methods. \textit{Full tuning} is not an efficient fine-tuning method but serves as a strong baseline for performance. \textit{BitFit~\cite{BenZaken2021BitFitSP}} tunes only bias terms of the model during training. \textit{Prompt-tuning~\cite{Lester2021ThePO}} inserts trainable prompt vectors to the inputs' embedding vectors. \textit{Adapters \cite{adapters}} appends a small residual MLP after every attention and feed-forward layer. We experiment with the sequential adapter version by \citet{adapters}. \textit{LoRA~\cite{lora}} inserts trainable low-rank matrices into each layer of the backbone Transformer model to parameterize the weights' changes. \textit{LST~\cite{lst}} hierarchically adds multiple side networks, with each side network responsible for modulating the activations of a specific layer in the pre-trained model. For all PETL methods and READ, we keep the backbone transformer frozen throughout the training and only update the new parameters.


\textbf{Datasets.}
We evaluate READ and the baselines on the GLUE \cite{Wang2018GLUEAM} benchmarks.
These benchmarks cover a variety of NLP tasks, including linguistic acceptability (CoLA \cite{Warstadt2018NeuralNA}), paraphrase detection (MRPC \cite{Dolan2005AutomaticallyCA}, QQP \cite{Chen2017QuoraQP}, STS-B \cite{Cer2017SemEval2017T1}), natural language inference (MNLI \cite{Williams2017ABC}, QNLI \cite{Rajpurkar2016SQuAD1Q}), 
and sentiment classification (SST-2)\footnote{We exclude RTE from GLUE due to its small size compared to other tasks}.
In GLUE, the original test set labels are not publicly available. Instead, we follow \cite{Zhang2020RevisitingFB} and \cite{compacter} to create a test set for each task as follows: if the training set contains less than 10k samples, we equally split the original validation set into two subsets and treat them as new validation and test sets; otherwise, we use the original validation set as the test set, and split 1k from the training set as the new validation set. For MNLI, we use the mismatched set as the validation set and the matched set as the test set. We report the dataset sizes in Appendix \ref{appendix:data}.

\textbf{Model Specification and Experimental Details.}
We adopt the encoder-decoder T5 \cite{Raffel2019ExploringTL} model as our backbone transformer. We use $\text{T5}_{\text{BASE}}$ for all of our experiments, and also use $\text{T5}_{\text{LARGE}}$ for READ experiments, which we denote by READ-large. We perform fine-tuning on each dataset for up to 30 epochs and do an early stop once validation metric stops improving. For READ, we experiment with $\{128, 256\}$ as RNN hidden dimensions, $\{\text{RNN, LSTM, GRU}\}$ as RNN architectures. For PETL baselines, we experiment with $\{32, 64\}$ as Adapters' bottleneck sizes, $\{8, 32\}$ as LoRA's ranks, and $\{10, 20, 30\}$ as Prompt-tuning's prompt sizes. For all experiments, we conduct a grid search for learning rates in between $[1\times 10^{-6}, 3\times 10^{-3}]$ on log scale for up to 32 rounds. We choose the hyperparameters that achieve the best validation scores and report their test scores. Complete setup and hyperparameters detail are in Appendix \ref{appendix:hp}.

\section{Evaluation Results}
\label{results_section}

We train and evaluate each method on all the GLUE tasks. We take the cumulative energy consumption and measure the peak GPU during training. In this section, we report and analyze the results on the GLUE Benchmarks. 

\textbf{READ outperforms other methods while consuming significantly lower energy:} 
Figure \ref{fig:combined_figures} (left) shows that READ can reduce GPU energy consumption by up to 90\% compared to full-tuning. READ lowers the GPU memory footprint by 56\% while retaining the same model accuracy when retraining. While other parameter-efficient transfer learning (PETL) methods like LoRA, BitFit or Adapter reduce the number of trainable parameters, they do not reduce the compute cost required to fine-tune. We believe the underlying optimization objective for PETL is to reduce this compute cost. Table \ref{tab:glue_results} shows the performance of all methods on GLUE with $\text{T5}{\text{BASE}}$. Excluding Adapter, READ outperforms all parameter-efficient methods while consuming at least 68\% less energy. Compared to Adapter, READ achieves nearly the same model accuracy (less than 0.1\% lower) while using 70\% less energy. More interestingly, READ with $\text{T5}{\text{LARGE}}$ (i.e. READ-large) achieves better performance than all other methods and consumes similar or less energy compared to other methods. For example, READ-large outperforms Full-tuning and Adapter by 1.4\% and 0.8\% with 69\% and 5\% less energy, respectively. These results show that by using READ, we can scale up the model size while keeping the same hardware and memory constraints.

\begin{table}[] 
\caption{Performance and energy consumption results of all methods on GLUE tasks. We report the accuracy for SST-2, MNLI, QNLI, and Matthew’s Correlation for CoLA. For STS-B we report the average of Pearson correlation and Spearman correlation. For MRPC and QQP, we report the average of F1 score and accuracy. For all tasks, we report the average score on 3 different seeds. Bold fonts indicate the best results of that column.}
\label{tab:glue_results}
\centering\small
\setlength\tabcolsep{2.9pt}
\begin{tabular}{l|c|cc|ccccccc|c}
\toprule
\multirow{2}{*}{Method} & \multirow{2}{*}{\begin{tabular}[c]{@{}c@{}}Trainable \\ Params (\%)\end{tabular}} & \multirow{2}{*}{\begin{tabular}[c]{@{}c@{}}Power\\     (kW)\end{tabular}} & \multirow{2}{*}{\begin{tabular}[c]{@{}c@{}}Energy \\    (kWh)\end{tabular}} & \multirow{2}{*}{CoLA} & \multirow{2}{*}{MNLI} & \multirow{2}{*}{QNLI} & \multirow{2}{*}{MRPC} & \multirow{2}{*}{QQP} & \multirow{2}{*}{SST-2} & \multirow{2}{*}{STS-B} & \multirow{2}{*}{Avg.} \\
                    &                                                                                   &                                                                               &                                                                                 &                       &                       &                       &                       &                      &                        &                        &                       \\
\midrule
\multicolumn{12}{c}{\cellcolor[HTML]{EFEFEF}\emph{\textbf{Baselines}}}\\     
\midrule
Full-tuning             & 100                                                                               & 0.77                                                                         & 12.52                                                                           & 53.97                 & 86.17                 & 90.87                 & 86.88                 & 89.71                & 92.89                  & 88.19                  & 84.52                 \\ 
Adapter                 & 0.96                                                                               & 0.50                                                                         & 6.99                                                                           & 52.56                 & 85.68                 & 92.89                 & 87.84                 & 88.95                & 93.12                  & 87.51                  & 85.04                 \\ 
LoRA                    & 0.48                                                                               & 0.68                                                                         & 10.58                                                                           & 51.70                 & 85.20                 & 92.72                 & \textbf{88.07}                 & 88.92                & 93.46                  & 86.97                  & 84.89                 \\ 
BitFit                  & 0.06                                                                               & 0.47                                                                         & 7.68                                                                           & 50.92                 & 85.28                 & 92.58                 & 86.32                 & 88.70                & \textbf{94.15}                  & 86.94                  & 84.43                 \\ 
Prompt-tuning           &  \textbf{0.01}                                                                             & 0.50                                                                         & 6.45                                                                           & 42.71                 & 79.38                 & 91.73                 & 86.04                 & 88.74                & 93.12                  & 84.96                  & 82.38                 \\
LST           & 2.00 & 0.44 & 10.59 & 53.38 & 84.53 & 92.43 & 87.38 & 88.31 & 92.09 & 87.37 & 84.58 \\
\midrule
\multicolumn{12}{c}{\cellcolor[HTML]{EFEFEF}\emph{\textbf{Our method}}}\\     
\midrule
READ                    & 0.80                                                                               & \textbf{0.43}                                                                         & \textbf{2.06}                                                                          & 52.59                 & 85.25                 & 92.93                 & 87.09                 & 89.10                & 93.80                  & 87.77                  & 84.97                 \\
READ-large              & 0.32                                                                               & 0.62                                                                         & 6.62                                                                           & \textbf{54.05}                 & \textbf{87.29}                 & \textbf{93.68}                 & 87.70                 & 89.34                & \textbf{93.92}                  & \textbf{88.58}                  & \textbf{85.73}              \\
\bottomrule
\end{tabular}
\end{table}


\begin{table}[]
\centering
\caption{READ with and without recurrency}
\label{read_recurrency}
\resizebox{\textwidth}{!}{
    \begin{tabular}{c
    >{\columncolor[HTML]{FFFFFF}}c
    >{\columncolor[HTML]{FFFFFF}}c
    >{\columncolor[HTML]{FFFFFF}}c
    >{\columncolor[HTML]{FFFFFF}}c 
    >{\columncolor[HTML]{FFFFFF}}c 
    >{\columncolor[HTML]{FFFFFF}}c 
    >{\columncolor[HTML]{FFFFFF}}c
    >{\columncolor[HTML]{FFFFFF}}c
    >{\columncolor[HTML]{FFFFFF}}c}

    \toprule
    Method & CoLA & MNLI & QNLI & MRPC & QQP & SST-2 & STS-B & Avg. & Trainable Params (\%) \\
    \midrule
    READ (with Recurrency) & 52.59 & 85.25 & 92.93 & 87.09 & 89.10 & 93.80 & 87.77 & 84.97 & 0.8 \\
    READ (w/o. Recurrency) & 53.24 & 84.10 & 91.51 & 89.02 & 89.18 & 94.04 & 87.10 & 85.15 & 9.6 \\
    READ-large (with Recurrency) & 54.05 & 87.29 & 93.68 & 87.70 & 89.34 & 93.92 & 88.58 & 85.73 & 0.32 \\
    READ-large (w/o Recurrency) & 50.17 & 86.90 & 93.00 & 87.61 & 89.12 & 94.15 & 87.46 & 85.02 & 6.4 \\
    \bottomrule
\end{tabular}
}
\end{table}

\begin{figure*}[ht]
    \centering
    \begin{minipage}[t]{0.495\linewidth}
        \centering
        \includegraphics[width=\linewidth]{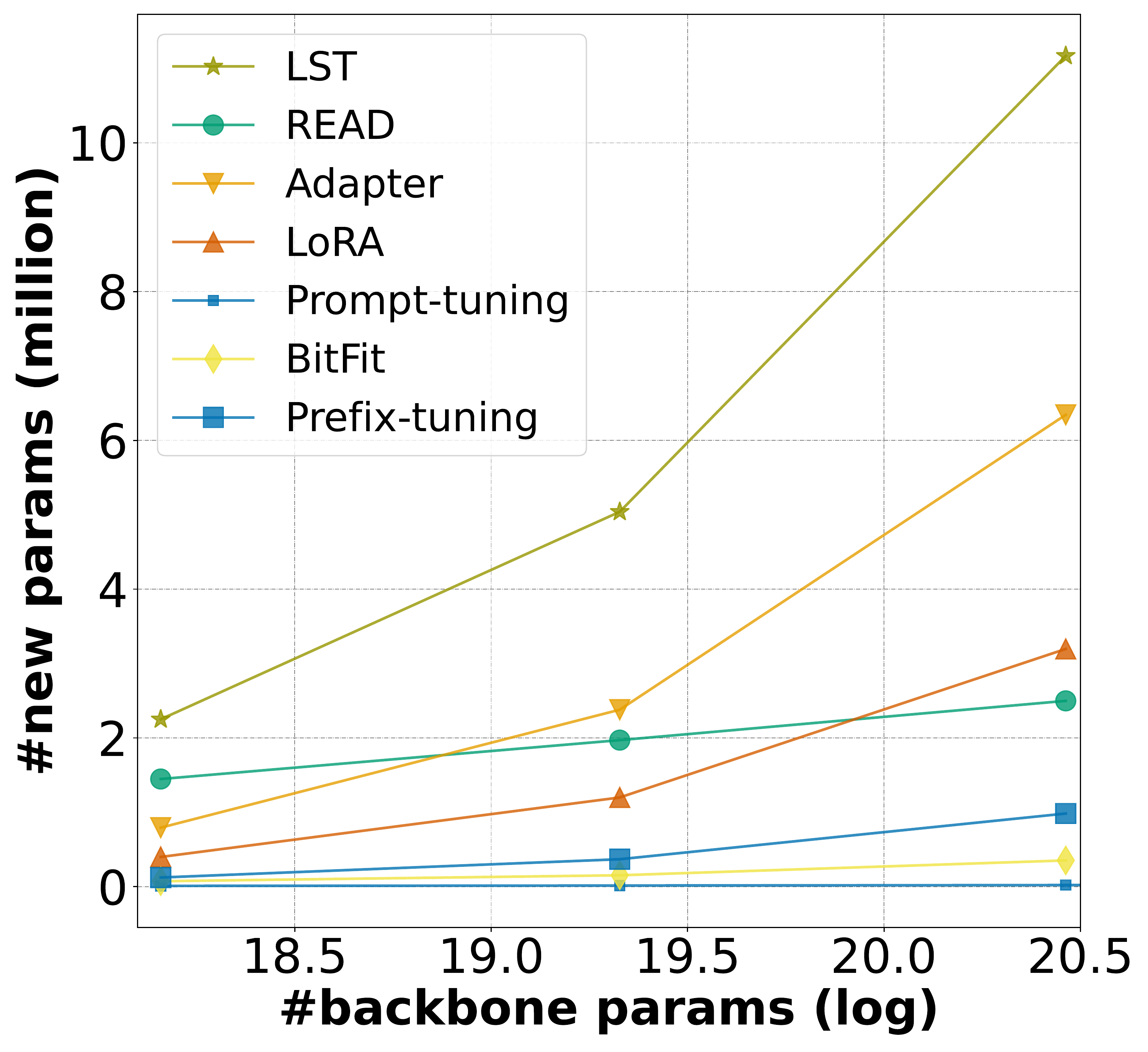}
        \caption{The number of trainable parameters as the backbone model size increases.}
        \label{adapted_size_backbone_size}
    \end{minipage}
    \hfill 
    \begin{minipage}[t]{0.495\linewidth}
        \centering
        \includegraphics[width=\linewidth]{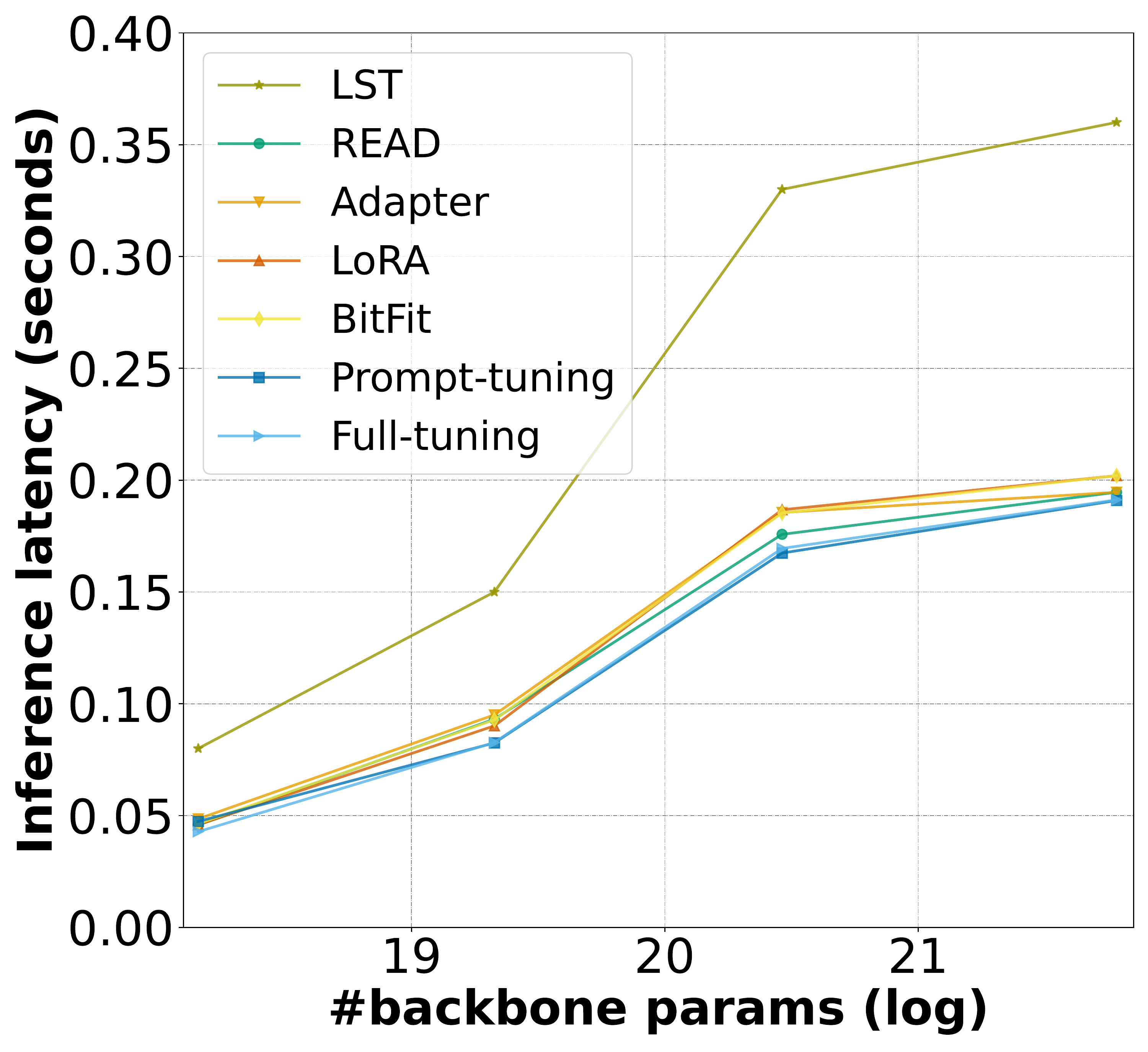}
        \caption{Inference latency as backbone model size increases.}
        \label{no_inference_decay}
    \end{minipage}
\end{figure*}

\textbf{READ consumes less training memory:} Figure \ref{fig:combined_figures} (right) shows the design space trade-off between  model quality performance and memory footprint. READ improves the training memory requirement by at least 25\% compared to all the other baselines while achieving similar or better performance. READ with $\text{T5}_{\text{LARGE}}$ consumes similar amount of memory as full-tuning with $\text{T5}_{\text{BASE}}$. As the backbone size increases, the memory savings achieved by READ become increasingly significant in comparison to the other PETL methods, as depicted in Figure \ref{adapted_size_backbone_size} (right). Notably, at the $\text{T5}_{\text{3B}}$ backbone level, these savings reach as high as 43\%. This observation suggests that READ is remarkably effective in the regime of fine tuning large Transformers.

\textbf{READ is scalable:} As shown in Figure \ref{adapted_size_backbone_size} (left), the number of trainable parameters of READ scale more slowly as compared to the other PETL methods. READ's number of parameters exhibits a log-linear growth pattern as the T5 backbone model size increases. In fact, the recurrent nature of READ makes its tunable size independent from the number of backbone layers, making READ a more suitable choice for fine-tuning large Transformers in practice.

\textbf{The importance of recurrency}
We perform ablation analysis on the importance of recurrence in READ in Table \ref{read_recurrency}. We find that the removal of recurrence does not significantly enhance READ quality and even diminishes quality for the T5 large backbone. However, without recurrence leads to over 12 times more trainable parameters, compromising scalability.

\textbf{Comparison with Ladder-Side-Tuning (LST)}
We compare our methods with Ladder-Side-Tuning (LST), another memory efficient fine-tuning approach \cite{lst}. We follow the pruning method introduced in \cite{lst} to extract a smaller transformer from the backbone transformer and use it to initialize the side transformer, and re-implement LST. Table \ref{tab:glue_results} lists the results of LST (using $\text{T5}_{\text{BASE}}$) on GLUE benchmarks and its energy efficiency.The results indicate that READ (base) outperforms LST (base) on most tasks (except for a tiny task MRPC), using $\boldsymbol{80\%}$ \textbf{less energy consumption} and $\boldsymbol{60\%}$ \textbf{less trainable parameters}. While LST consumes $15\%$ less peak training memory relative to READ, it takes $\boldsymbol{40\%}$ \textbf{more inference memory} and $\boldsymbol{77\%}$ \textbf{longer inference time} than READ, a consequence of its attention-based side-network architecture. It is also noteworthy that when compared to LST even READ-large saves $38\%$ GPU energy and yields a similar inference latency, with $1.4\%$ relative gain on the averaged GLUE score. Furthermore, the "pre-training stage" refers to the process described in LST paper section 2.2, where distillation is performed with T5 pre-training objective. It is important to note that caching the attention outputs does not involve updating any model parameters and should not be considered as a form of training.

\section{Related Work}
In this section, we summarize the closely related works to ours and leave the more detailed discussion to Appendix \ref{appendix:related_works}.

\textbf{Parameter-efficient Transfer Learning.}
The surge in generative AI applications \cite{stable_diffusion, flan, llama, biderman2023pythia} has been hindered by the computational and memory costs of fine-tuning large transformers. Parameter-efficient transfer learning (PETL) \cite{bitfit, adapters, aghajanyan2020intrinsic, prefix, prompt, petl_survey, lst} addresses this by training a minimal parameter set, with various methods like Adapter-based approaches, Low-Rank Adaptation (LoRA), BitFit, and Prompt-tuning. Unlike these, READ introduces memory efficiency by incorporating a small recurrent network into the backbone, focusing on reducing memory usage over parameter minimization.

\textbf{Memory-Efficient Training.}
Memory-efficient training strategies, such as gradient checkpointing \cite{grad_check}, reversible layers \cite{reversible}, ZeRO \cite{zero}, and Layer-wise Adaptive Rate Scaling (LARS) \cite{lars}, aim to mitigate memory consumption by optimizing the storage and computation of intermediate activations and model states, particularly in distributed training environments.

\textbf{Sidenet Tuning.}
Side-tuning \cite{side_tuning} and Ladder side-tuning \cite{lst} employ lightweight side networks to adapt pre-trained model activations for new tasks without modifying the base model. READ, while inspired by these, distinguishes itself by utilizing a single RNN block that processes the backbone network's hidden state recurrently, ensuring the fine-tuning parameter count does not scale with the backbone size. Unlike Side-Tuning, READ iteratively calculates its RNN states across all layers and exclusively employs RNN and Feed-Forward Network (FNN) structures, negating the need for transformers or attention mechanisms and enabling use without pre-training.

\section{Conclusion and Limitations}
\textbf{Limitations.} Due to our limited computing resources, we could not scale the backbone to an even larger scale. A future direction is to fine-tune READ on Llama-7B \cite{llama} or even larger variants. Another direction can be studied if READ can generalize well in a low-data regime. A drawback of READ is its tendency to require more epochs to converge on small datasets than other PETL methods. Consequently, although READ is more efficient in per-unit time computations, it may not yield significant overall consumption gains when a task has few data points. We leave investigating READ on the low-data regime as future work. 

\textbf{Conclusion.}
In this paper, we propose REcurrent ADaption (READ), a lightweight and efficient parameter and memory-efficient fine-tuning method, for large-scale transformers. We show that READ achieves comparable accuracy to full fine-tuning while saving more than 84\%  of energy consumption and reducing training memory consumption by 56\% relative to full-tuning. We demonstrate the scalability of READ because READ is independent of backbone model size. We hope that READ can make fine-tuning large models more accessible to a broader range of researchers and applications.

\bibliography{reference}
\bibliographystyle{plainnat}

\newpage

\appendix
\section{Appendix}
\label{readderive}
\subsection{Revisit Transformer}
\label{notation}
In this subsection, we briefly revisit the computation of transformer and introduce some convenient notations for the future. Let $\mathcal{T}$ be a transformer of dimension $d$ with $N$ layers $\mathcal{L}_1,\cdots, \mathcal{L}_N$. At each layer $\mathcal{L}_i$, let the feed-forward network be $\mathcal{F}_i$ and multihead-attention be $\mathcal{A}_i$. Given a context sequence of $m$ tokens, we can express each layer as a mapping from $\mathbb{R}^{m\times d}$ to $\mathbb{R}^{m\times d}$ as follows:
\begin{equation}
\label{layer}
\mathcal{L}_i= (\mathcal{F}_i^* + I)\circ \mathcal{A}_i^* + I,
\end{equation}
where $I$ represents the identity mapping, $\circ$ denotes mapping composition, and $\mathcal{F}_i^* \eqqcolon \mathcal{F}_i\circ \text{LN}$, $\mathcal{A}_i^* \eqqcolon \mathcal{A}_i\circ \text{LN}$ (i.e. compositions with layer-normalization).
Further, we define $\mathcal{R}_i = (\mathcal{F}_i^* + I)\circ \mathcal{A}_i^*$ so as to write layer mapping as $\mathcal{L}_i = \mathcal{R}_i + I.$


\subsection{Derivations of READ}
Following the notations in Subsection \ref{notation}, we derive an inductive formula for the \emph{corrections}:
\begin{eqnarray}
\begin{split}
\label{deltainduction}
\delta\phi_{i} &= \phi_{i}^\prime -  \phi_{i} \\
&= (\mathcal{R}^\prime_i + I)(\phi^\prime_{i-1}) - (\mathcal{R}_i + I)(\phi_{i-1}) \\
&= \mathcal{R}^\prime_i(\phi^\prime_{i-1}) - \mathcal{R}_i(\phi_{i-1}) + (\phi^\prime_{i-1} - \phi_{i-1})\\
&= \mathcal{R}^\prime_i(\phi^\prime_{i-1}) - \mathcal{R}_i(\phi_{i-1}) + \delta\phi_{i-1}\\
&= \big{(}\mathcal{R}^\prime_{i-1} - \mathcal{R}_i\big{)}(\phi^\prime_{i-1}) + \big{(}\mathcal{R}_i(\phi^\prime_{i-1}) - \mathcal{R}_i(\phi_{i-1})\big{)} + \delta\phi_{i-1}\\
&= \delta\mathcal{R}_i(\phi_{i-1}^\prime) + J\mathcal{R}_i\delta\phi_{i-1} + \delta\phi_{i-1}.
\end{split}
\end{eqnarray}
Here $\delta\mathcal{R}_i$ denotes the operator difference $\mathcal{R}^\prime_i - \mathcal{R}_i$, and $J\mathcal{R}_i$ is the Jacobian matrix of $\mathcal{R}_i(\cdot)$ evaluated at some point lying on the line segment from $\phi_{i-1}$ to $\phi_{i-1}^\prime$. To simplify our arguments, we (1) assume that $J\mathcal{R}_i$ takes value at $\phi_{i-1}$, (2) let $\mathcal{T}^\prime$ be the consequence of fine-tuning with Adapter or LoRA (applied at FFN layers $\mathcal{F}_i$) \footnote{For fine-tuning methods that modify attention, we expect a similar conclusion that demands a more intricate line of reasoning, which we defer to future research.}. We use $\mathcal{P}$ to denote a common module adopted by Adapter and LoRA which consists of a down projection matrix to a lower dimension possibly followed by a non-linear activation, and then composed with an upper projection back to the original dimension\footnote{The operator norm of $\mathcal{P}$ is small when its two matrices have small weights, and therefore addition with $\mathcal{P}$ will not change the invertibility of an already invertible operator.}. Under these assumptions, the first term of the RHS in \eqref{deltainduction} now becomes
\begin{eqnarray}
\begin{split}
\label{deltar}
    \delta\mathcal{R}_{i}(\phi_{i-1}^\prime) &=\begin{cases}
			\mathcal{P}_i \circ (\mathcal{P}_i + I)^{-1}\phi_{i}^\prime & \text{(Adapter)}\\
            \mathcal{P}_i \circ (\mathcal{P}_i +\mathcal{F}_i)^{-1}\phi_{i}^\prime & \text{(LoRA)}
		 \end{cases}\\
   &\eqqcolon \mathcal{W}_i(\phi_{i} + \delta\phi_{i})
\end{split}
\end{eqnarray}
Now plugging \eqref{deltar} back to \eqref{deltainduction}, we obtain
\begin{equation}
\label{adapterdeltapre}
\delta\phi_{i} = \mathcal{W}_i(\phi_{i} + \delta\phi_i) + J\mathcal{R}_i\delta\phi_{i-1} + \delta\phi_{i-1}.
\end{equation}
Notice that both sides of equation \eqref{adapterdeltapre} contains $\delta\phi_i$. Because of the non-linearity of $W_i$, there is no straightforward way to extract an inductive formula of $\delta\phi_i$ from \eqref{adapterdeltapre}. 

However, let us rewrite equation \eqref{adapterdeltapre} as 
\begin{eqnarray}
\begin{split}
&\delta\phi_{i} - \mathcal{W}_i(\phi_{i} + \delta\phi_i) - (J\mathcal{R}_i\delta\phi_{i-1} + \delta\phi_{i-1}) \\
=& F(\delta\phi_i, \phi_{i}, J\mathcal{R}_i\delta\phi_{i-1} + \delta\phi_{i-1}) =0,
\end{split}
\end{eqnarray}
and compute the Jacobian to see that $J_{\delta\phi_i}F = I - J\mathcal{W}_i$, which is invertible when $\mathcal{P}_i$ (and hence $\mathcal{W}_i$ \ref{deltar}) has small norm. Now by Implicit Function Theorem there exists $G$ such that
\begin{equation}
\delta\phi_i = G(\phi_{i}, J\mathcal{R}_i\delta\phi_{i-1} + \delta\phi_{i-1}).
\end{equation}
An alternative argument is to use a first order approximation of $\mathcal{W}_i(\phi_i + \delta\phi_i)$ assuming that $\delta\phi_i$ is sufficiently small, which gives us the following inductive formula:  
\begin{equation}
\label{adapterdelta}
\delta\phi_{i} = (I-J\mathcal{W}_i)^{-1}\circ \bigg{(} \mathcal{W}_i\phi_{i} + J\mathcal{R}_i\delta\phi_{i-1} + \delta\phi_{i-1}\bigg{)}
\end{equation}
We take the second approach above and adopt formula \eqref{adapterdelta} as we move forward, because of its explicit function form.
Note that every operation in \eqref{adapterdelta} acts on the column space of $\phi$ except for the Jacobian transform $J\mathcal{R}_i$, so let us first focus on expanding $J\mathcal{R}_i\delta\phi_{i-1}$. In fact, we will compute the Jacobian for a general attention mapping that takes $3$ arguments $\phi_q, \phi_k, \phi_v$ (i.e. hidden states of queries, keys, and values), and then apply the results to the special case of self-attention (as in encoder) and cross-attention (as in decoder) respectively. For the sake of brevity, we assume that the number of attention head is $1$ and omit the output projection, as neither of which is essential to our conclusion.  

Let $\phi_q, \phi_k, \phi_v$ be matrices in $\mathbb{R}^{m_q\times d},\mathbb{R}^{m_k\times d},\mathbb{R}^{m_k\times d}$, which stand for $\mathbb{R}^d$-vector representations of the query, key, and value token sequences with length $m_q, m_k, m_k$ respectively. We use an upper index $\alpha$ to denote the vector associated to the $\alpha^{\text{th}}$ token, and omit the lower layer index $i$ when no ambiguity is present (e.g. $\mathcal{A}^\alpha$ is the $\alpha^{\text{th}}$ column of $\mathcal{A}$'s output.). First, we have 
\begin{equation}
\label{chain}
J\mathcal{R}_i\delta\phi_{i-1}  = (J\mathcal{F}^* + I)\circ J\mathcal{A}\circ J\text{LN}(\delta\phi_{i-1})
\end{equation}
by chain rule. Next we expand $J\mathcal{A}$, as every other operation in \eqref{chain} acts on the column space of $\phi$; especially, up to composing with a feed-forward neural network, let us replace $J\text{LN}$ by an identity to simplify our notations. A straightforward computation gives the following:
\begin{eqnarray}
\begin{split}
\label{attnjac}
J_{ \phi_q}\mathcal{A}^\alpha(\delta\phi_q) &=\bigg{[}\sum^{m_q}_{\beta=1}\sigma^{\alpha\beta}\cdot \frac{(v^\beta - \mathcal{A}^\alpha) \cdot {k^\beta}^T}{\sqrt{d}}\cdot W_Q\bigg{]} \delta \phi_q^\alpha,\\
J_{ \phi_k}\mathcal{A}^\alpha(\delta\phi_k)
&= \sum^{m_k}_{\beta=1}\sigma^{\alpha\beta}\cdot \frac{(v^\beta - \mathcal{A}^\alpha) \cdot {q^\alpha}^T}{\sqrt{d}}\cdot W_K \delta \phi_k^\beta,\\
J_{ \phi_v}\mathcal{A}^\alpha(\delta\phi_v)&= \sum_{\beta=1}^{m_k} \sigma^{\alpha\beta}\cdot W_V\delta \phi_v^\beta.
\end{split}
\end{eqnarray}
Here $W_Q, W_K, W_V$ denote the query, key, and value projection matrices of $\mathcal{A}$, $q^\alpha=W_Q\phi_q^\alpha$, $k^\alpha=W_K\phi_k^\alpha$, and $\sigma^{\alpha\beta}=\text{softmax}({q^\alpha}^T\cdot k^\beta/\sqrt{d})$.

\textbf{Case 1, $\mathcal{A}$ is self-attention} Upon setting $\phi_q, \phi_k, \phi_v$ to $\phi$, and $\delta \phi_q, \delta \phi_k, \delta \phi_v$ to $\delta\phi$ in \eqref{attnjac}, we obtain:
\begin{eqnarray}
\begin{split}
\label{selfattnjac}
J\mathcal{A}^\alpha \delta\phi =& \bigg{[}\sum^{m_q}_{\beta=1}\sigma^{\alpha\beta}\cdot \frac{(v^\beta - \mathcal{A}^\alpha) \cdot {k^\beta}^T}{\sqrt{d}}\cdot W_Q\bigg{]} \delta\phi^\alpha\\
&+ \sum^{m_q}_{\beta=1}\sigma^{\alpha\beta}\cdot \bigg{[}\frac{(v^\beta - \mathcal{A}^\alpha) \cdot {q^\alpha}^T}{\sqrt{d}}\cdot W_K + W_V\bigg{]}\delta\phi^\beta. 
\end{split}
\end{eqnarray}
Note the two quantities in the square brackets are $\mathbb{R}^{d\times d}$-matrix-valued linear functions of values that can be computed from the cached results at $\mathcal{L}_i$, which we shall denote by $\Phi, \Psi$ from now on:
\begin{equation}
\label{jacenc}
J\mathcal{A}^\alpha \delta\phi = \Phi \cdot \delta\phi^\alpha + \sum_{\beta=1}^{m_q} \sigma^{\alpha\beta}\Psi \cdot\delta\phi^\beta.
\end{equation}
Now, upon inserting \eqref{jacenc} to the $\alpha^{\text{th}}$-column of \eqref{chain} by setting $\phi$ as $\phi_i$, and then plugging \eqref{chain} back in \eqref{adapterdelta}, we obtain the iterative formula for outputs $h_i$:
\begin{eqnarray}
\begin{split}
\label{protoreadpost}
\begin{cases}
            \psi_i^\alpha = \Phi_i\cdot \mathcal{F}_i\delta\phi_{i-1}^\alpha + \sum_{\beta=1}^{m}\sigma_i^{\alpha\beta} \Psi_i\cdot \mathcal{F}_i\delta\phi_{i-1}^\beta\\[5pt]
			x_{i}^\alpha = [{\phi^\alpha_i}^T,{\psi_i^\alpha}^T ]^T\\[5pt]
            \delta\phi_{i}^\alpha = \mathcal{G}_i(\mathcal{H}_ix_i^\alpha + \delta\phi_{i-1}^\alpha)
		 \end{cases}
\end{split}
\end{eqnarray}
where $\Phi_i, \Psi_i$ are defined as in \eqref{jacenc}, and $\mathcal{F}_i, \mathcal{G}_i, \mathcal{H}_i$ are FNNs that simulate $J\text{LN}$, $(I-J\mathcal{W}_i)^{-1}$, and $[\mathcal{W}_i, J\mathcal{F}^*_i + I]$ respectively; see \eqref{adapterdelta} and \eqref{chain}. Note \eqref{protoreadpost} is exactly \eqref{protoread} upon replacing $\delta\phi$ by $h$.

\textbf{Case 2, $\mathcal{A}$ is cross-attention} Since the decoder's \emph{correction} iterative formula follows from a similar line of reasoning as self attention, we present the final results while omitting the details:
\begin{eqnarray}
\begin{split}
\label{protoreadpostdec}
\begin{cases}
            \psi_i^\alpha = \Phi_i\cdot \mathcal{F}^D_i \delta\phi_{i-1}^{D,\alpha} + \sum_{\beta=1}^{m}\sigma_i^{\alpha\beta} \Psi_i\cdot \mathcal{F}^E_i \delta\phi^{E,\beta}\\[5pt]
			x_{i}^\alpha = [{\phi^\alpha_i}^T,{\psi_i^\alpha}^T ]^T\\[5pt]
            \delta\phi_{i}^{D,\alpha} = \mathcal{G}_i(\mathcal{H}_ix_i^\alpha + \delta\phi_{i-1}^{D,\alpha})
		 \end{cases}
\end{split}
\end{eqnarray}
where an upper index $D\backslash E$ are used to distinguish between the hidden states of decoder and encoder, and $\delta\phi^E$ is the final \emph{correction} of encoder.

\subsection{Ablation Experiments}
\label{appendix:abalation}
\begin{table}[]\centering
\caption{Efficiency results of LST, READ, and Full-tuning. We report the training GPU energy usage summed over all tasks, and the peak training memory (per batch) averaged over all tasks. For inference memory/time, we use MNLI and report the average per batch (with test batch size $1$).}\label{lst_other_results}
\setlength\tabcolsep{3.9pt}
\begin{tabular}{
>{\columncolor[HTML]{FFFFFF}}c |
>{\columncolor[HTML]{FFFFFF}}c 
>{\columncolor[HTML]{FFFFFF}}c 
>{\columncolor[HTML]{FFFFFF}}c 
>{\columncolor[HTML]{FFFFFF}}c 
>{\columncolor[HTML]{FFFFFF}}c }
\toprule
           & \begin{tabular}[c]{@{}c@{}}Training GPU \\ Energy (kWh)\end{tabular} & \begin{tabular}[c]{@{}c@{}}Training \\ Memory (GB)\end{tabular} & \begin{tabular}[c]{@{}c@{}}Trainable Params\\         (Million/\%)\end{tabular} & \begin{tabular}[c]{@{}c@{}}Inference \\  Time (s)\end{tabular} & \begin{tabular}[c]{@{}c@{}}Inference\\ Memory (GB)\end{tabular} \\ \midrule
Full-tuning & 12.52                                                                & 17.86                                                           & 247.58/100.00                                                                 & \textbf{0.083}                                                      & \textbf{0.943}                                                         \\
LST        & 10.59                                                                 & \textbf{5.77}                                                            & 5.04/2.00                                                                 & 0.165                                                          & 1.358                                                           \\
READ       & \textbf{2.07}                                                                 & 6.90                                                            & \textbf{1.97/0.80}                                                                 & 0.093                                                          & 0.966                                                           \\
READ-large & 6.62                                                                & 17.74                                                           & 11.17/1.4                                                                 & 0.175                                                          & 2.943                                                           \\ \bottomrule
\end{tabular}
\end{table}

\subsection{GPU Energy Analysis}
To provide a comprehensive understanding, we include an analysis below to show the mean and standard deviation for the sums of GPU energy, epochs to convergence, and training time across all GLUE tasks. While this analysis does reveal some variations in energy/time levels, they are not significantly substantial to alter the general trend, as READ continues to stand out as the most energy-efficient approach, with faster convergence than most baselines except for full-tuning. 

\begin{table}[h]
\centering
\begin{tabular}{|c|c|c|c|c|c|c|c|}
\hline
 & Full  & Adapter  & LoRA & Prompt & BitFit & LST & READ \\
\hline
Energy & $12.52_{0.44}$ & $6.99_{0.62}$ & $10.58_{1.9}$ & $6.45_{0.24}$ & $7.68_{0.06 }$ & $10.59_{0.3}$ & $2.06_{0.18}$ \\
\hline
Epoch & $7.0_{0}$ & $13.01_{0.58}$ & $25.84_{3.9}$ & $34.85_{2.4}$ & $23.61_{1.3}$ & $23.58_{1.5}$ & $12.31_{1.4}$ \\
\hline
Time & $472.74_{12.77}$ & $485.2_{51.12}$ & $409.4_{19.52}$ & $315.53_{5.01}$ & $292.89_{4.03}$ & $984.55_{19.07}$ & $155.11_{8.4}$ \\
\hline
\end{tabular}
\caption{GPU Energy Consumption and Training Time across 3 trials.}
\end{table}

\subsection{READ's Inference and Memory Efficiency}
\begin{table}[]\centering
\caption{Average inference memory consumption (GB) for every method with different backbones on GLUE benchmark.}\label{inf_mem_table}
\begin{tabular}{c
>{\columncolor[HTML]{FFFFFF}}c
>{\columncolor[HTML]{FFFFFF}}c
>{\columncolor[HTML]{FFFFFF}}c
>{\columncolor[HTML]{FFFFFF}}c 
>{\columncolor[HTML]{FFFFFF}}c 
>{\columncolor[HTML]{FFFFFF}}c 
>{\columncolor[HTML]{FFFFFF}}c}
\toprule
                              & READ    & LST & Adapter  & LoRA     & Prompt   & Bias     & Full     \\ \midrule
\cellcolor[HTML]{FFFFFF}$\text{T5}_{\text{SMALL}}$ & 0.317  & 0.427 & 0.303  & 0.302  & 0.301  & 0.301  & 0.300  \\
\cellcolor[HTML]{FFFFFF}$\text{T5}_{\text{BASE}}$  & 0.966  & 1.358 & 0.952  & 0.948  & 0.948   & 0.945  & 0.943  \\
\cellcolor[HTML]{FFFFFF}$\text{T5}_{\text{LARGE}}$ & 2.943  & 4.597 & 2.936  & 2.925   & 2.925  & 2.914  & 2.912  \\
\cellcolor[HTML]{FFFFFF}$\text{T5}_{\text{3B}}$    & 10.885 & 11.400 & 10.878 & 10.866 & 10.894 & 10.855 & 10.853 \\ \bottomrule
\end{tabular}
\end{table}
\label{sec:inference}
As Figure \ref{no_inference_decay} (left) and Table \ref{inf_mem_table} indicate, READ achieves comparable inference latency and memory requirement as the other PETL methods. To assess the inference memory impact of READ more comprehensively, we use Figure \ref{no_inference_decay} (right) to demonstrate that, as the backbone size increases, the inference memory growth (relative to Full-tuning) of READ becomes less noticeable and decays to a similar extent as the other methods at $\text{T5}_{\text{LARGE}}$. 

\begin{table}[]\centering
\caption{Split sizes, training GPU number, and training batch size per GPU node for all GLUE tasks.}\label{data_size}
\begin{tabular}{c
>{\columncolor[HTML]{FFFFFF}}c 
>{\columncolor[HTML]{FFFFFF}}c 
>{\columncolor[HTML]{FFFFFF}}c 
>{\columncolor[HTML]{FFFFFF}}c 
>{\columncolor[HTML]{FFFFFF}}c 
>{\columncolor[HTML]{FFFFFF}}c 
>{\columncolor[HTML]{FFFFFF}}c }
\toprule
\cellcolor[HTML]{FFFFFF}{\color[HTML]{000000} }                                                                  & {\color[HTML]{000000} CoLA} & {\color[HTML]{000000} MNLI}  & {\color[HTML]{000000} QNLI} & {\color[HTML]{000000} MRPC} & {\color[HTML]{000000} QQP}   & {\color[HTML]{000000} SST-2} & {\color[HTML]{000000} STS-B} \\ \midrule
\cellcolor[HTML]{FFFFFF}{\color[HTML]{000000} \begin{tabular}[c]{@{}c@{}}Training\\ Samples (k)\end{tabular}}    & {\color[HTML]{000000} 8.5}  & {\color[HTML]{000000} 392.7} & {\color[HTML]{000000} 99.3} & {\color[HTML]{000000} 3.7}  & {\color[HTML]{000000} 323.4} & {\color[HTML]{000000} 66.5}  & {\color[HTML]{000000} 5.8}   \\ \midrule
\cellcolor[HTML]{FFFFFF}{\color[HTML]{000000} \begin{tabular}[c]{@{}c@{}}Test\\ Samples (k)\end{tabular}}        & {\color[HTML]{000000} 0.52} & {\color[HTML]{000000} 9.8}   & {\color[HTML]{000000} 5.4}  & {\color[HTML]{000000} 0.2}  & {\color[HTML]{000000} 40.4}  & {\color[HTML]{000000} 0.9}   & {\color[HTML]{000000} 0.8}   \\ \midrule
\cellcolor[HTML]{FFFFFF}{\color[HTML]{000000} \begin{tabular}[c]{@{}c@{}}Validation \\ Samples (k)\end{tabular}} & {\color[HTML]{000000} 0.52} & {\color[HTML]{000000} 9.8}   & {\color[HTML]{000000} 1.0}  & {\color[HTML]{000000} 0.2}  & {\color[HTML]{000000} 1.0}   & {\color[HTML]{000000} 1.0}   & {\color[HTML]{000000} 0.8}   \\ \midrule
GPUs                                                                                                             & 2                           & 8                            & 8                           & 1                           & 8                            & 8                            & 1                            \\ \midrule
\begin{tabular}[c]{@{}c@{}}Batch Size\\  per GPU\end{tabular}                                        & 48                          & 12                           & 12                          & 96                          & 12                           & 12                           & 96                           \\ \bottomrule
\end{tabular}
\end{table}

\section{Architecture}
\label{networkchoices}
\subsection{Architecture choices}
The matrix functions $\Psi, \Phi$ in equation \eqref{protoreadpost} and \eqref{protoreadpostdec} requires computing dot products for $m^2$ pairs of vectors \eqref{attnjac} with time complexity as large as $O(m^2d^2)$. To reduce latency cost in practice, we make substantial reductions to the first equation in both \eqref{protoreadpost} and \eqref{protoreadpostdec} for READ experiments in this paper, as listed below:
\begin{itemize}
    \item Indices $i$s are removed and learnable parameters are fused across all layers;
    \item In self-attention, we set $\Psi, \Phi$ to be constantly zeros; in other words, only hidden states are cached and used for encoder \emph{corrections};
    \item In cross-attention, we set $\Phi$ to zero and $\Psi\cdot \mathcal{F}^E_i h_{i-1}^\beta\eqqcolon Lh_{i-1}^\beta$, where $L$ is a learnable linear projection, so besides decoder hidden states we also need to cache the cross-attention scores for computing decoder \emph{corrections}. Furthermore, we use a simple addition operation to combine $\phi_i$ and $\psi_i$ in \eqref{protoreadpostdec} instead of a learnable layer.
\end{itemize}
Note some reductions we made above might be over-simplified but this paper does not explore other more sophisticated\footnote{A more sophisticated choice potentially introduces more dependency on cached results and likely to improve performance at a trade-off of higher number of computation flops.} while still computationally efficient options, such as a gated neural network:
\begin{eqnarray}
\begin{split}
\begin{cases}
\Phi\cdot \mathcal{F}_i(h_{i-1}^\alpha)=\text{Gate}(\phi^\alpha_{i})\odot \text{FFN}(h^\alpha_{i-1}),\\[3pt]
\Psi\cdot \mathcal{F}_i(h_{i-1}^\beta)=\text{Gate}(\phi^\alpha_{i})\odot \text{FFN}(h^\beta_{i-1}),
		 \end{cases}
\end{split}
\end{eqnarray}
where $v\odot X = \text{diag}(v)\cdot X$. We leave the pertinent explorations to future works.

\subsection{READ Algorithm}
\label{algosection}
Algorithm \ref{readalgo} outlines a forward pass during READ fine-tuning. Let $\mathcal{T}$ be a transformer with $N^E$ encoder layers and $N^D$ decoder layers, and $X\backslash Y$ be source$\backslash$target sequences of length $m\backslash n$:
\begin{algorithm}
\setstretch{1.8}
\caption{READ Fine Tuning Algorithm}\label{readalgo}
\begin{algorithmic}
\State Initialize RNNs $\mathcal{N}^E$, $\mathcal{N}^D$ and a learnable projection $\Psi$.
\State $\{\phi_{i}^{E,\alpha}\}_{i=1, \alpha=1}^{N^E, m}, \{\phi_{j}^{D,\alpha}\}_{j=1, \alpha=1}^{N^D, n}, \{\sigma_{i}^{E,\alpha\beta}\}_{i=1, \alpha=1, \beta=1}^{N^E, m, m}, \{\sigma_{j}^{D,\alpha\beta}\}_{j=1, \alpha=1, \beta=1}^{N^D, n, m} \gets \mathcal{T}(X, Y)$ 
\State $h_{E,0}\gets 0$ \textcolor{blue}{\Comment{We assume embeddings need no \emph{corrections}}}.
\For{$i$ in $1,\cdots, N^E$} \textcolor{blue}{\Comment{{Iteratively compute encoder \emph{corrections}}}}. 
    \State $h_{i}^{E, \alpha} = \mathcal{N}^E(\phi_i^\alpha, h_{i-1}^{E,\alpha})$, $\forall\alpha$
\EndFor
\State $h_{D,0}\gets 0$ 
\For{$j$ in $1,\cdots, N_D$} \textcolor{blue}{\Comment{{Iteratively compute decoder \emph{corrections}}}}. 
    \State $\psi_{j}^\alpha = \sum_{\beta=1}^{m}\sigma_{D,j}^{\alpha\beta}\Psi h_{N^E}^{E,\beta}$, $\forall\alpha$
    \State $x_{j}^\alpha = \phi_j^\alpha + \psi_j^\alpha$ , $\forall\alpha$
    \State $h_{ j}^{D,\alpha} = \mathcal{N}^D(x_j^\alpha, h_{ j-1}^{D,\alpha})$, $\forall\alpha$
\EndFor
\State ${\phi_{ N_D}^D}^\prime \gets \phi_{ N^D}^D + h_{N^D}^D$ \textcolor{blue}{\Comment{Obtain \emph{adapted} outputs}}.
\end{algorithmic}
\end{algorithm}

\section{Related Works}
\label{appendix:related_works}
\textbf{Parameter-efficient Transfer Learning.}
There has been an explosion of generative AI applications in recent months \cite{stable_diffusion, flan, llama, biderman2023pythia}. However, the ability to fine-tune large transformers is primarily limited by the growing compute cost required to adapt and serve these models. Parameter-efficient transfer learning (PETL) \cite{bitfit, adapters, aghajanyan2020intrinsic, prefix, prompt, petl_survey, lst} aims to solve this problem by training only a small set of parameters. There are many PETL methods which we defer the reader to \cite{petl_survey} for a more comprehensive overview. In this section, we will summarize the most popular PETL methods which we used as baselines. Adapter-based approaches \cite{adapters, parallel_adapters} insert small learnable modules between pre-trained model layers and only update these adapters during fine-tuning, reducing computational cost and memory requirements. Low-Rank Adaptation (LoRA) \cite{lora} injects trainable rank decomposition matrices into each layer of the Transformer model. BitFit \cite{bitfit} fine-tunes only the biases of the model. Prompt-tuning \cite{prompt} is a successor of Prefix-Tuning \cite{prefix}, which adds a continuous task-specific prompt to the input.
In contrast, current PETL approaches aim to minimize the number of parameters trained. These approaches do not lead to memory efficiency, a more meaningful objective than parameter efficiency. This work proposes READ, simple memory-efficient methods by inserting a small recurrent network into the backbone. 

\textbf{Memory-Efficient Training.}
Memory-efficient training reduces memory consumption by reducing the storage of intermediate activations \cite{lst}. Gradient checkpointing \cite{grad_check} reduces memory consumption during backpropagation by storing a subset of intermediate activations and recomputing them as needed, trading time for memory. Reversible layers \cite{reversible} reconstruct each layer's activations from the next layer's activations. ZeRO \cite{zero} partitions model states, gradients, and optimizer states across multiple devices for distributed training, significantly reducing memory redundancy. Layer-wise Adaptive Rate Scaling (LARS) \cite{lars} dynamically scales learning rates for different layers, reducing memory overhead associated with large gradients and enabling the training of large models with limited memory.

\textbf{Sidenet Tuning.}
Side-tuning \cite{side_tuning} adds a lightweight side network alongside the pre-trained model. During training, the side network and the task-specific head are updated while the pre-trained model's parameters are kept fixed. The side network learns to modulate the pre-trained model's activations, allowing it to adapt to the new task without altering the base model. Ladder side-tuning \cite{lst} hierarchically adds multiple side networks, with each side network responsible for modulating the activations of a specific layer in the pre-trained model. While READ takes inspiration from Side-Tuning and LST, we would like to highlight significant differences between READ and prior works. First, READ only contains a single RNN block which takes in the hidden state of the backbone network in a recurrent manner. This way, the number of parameters to fine-tune does not increase with the size of the backbone, whereas LST attaches multiple transformer blocks to the backbone network. When the backbone gets larger, the size of the LST network also gets larger. Secondly, Side-Tuning uses an additive side network to sum its representation with the backbone network in only the last layer. READ consumes the backbone's hidden state at every layer to iteratively calculate its RNN states. The recurrence nature of RNN allows information to flow from one layer to the next, which is why READ outperforms other PETL methods. Last, our fine-tuning is transformer-free as only RNN and Feed-Forward Network (FNN) structures are used in READ and require no transformer or attention mechanism. We may use a randomly initialized READ network without going through pre-training like in LST or exploiting any subtle tricks for training a transformer.

\section{Experimental Details}
\label{experiment_details}
\textbf{Energy Consumption Measurement.}
\label{energy}
Higher training efficiency translates to lower energy consumption. 
To demonstrate the training efficiency benefit of READ, we measure and report the training GPU energy consumption (in kWh) for every experiment. We adopt the following commonly-used methodology to measure and estimate the model training energy consumption. We take the GPU resource utilization into account when computing the corresponding energy consumption by assuming a simple linear relationship between GPU utilization and its power consumption. Assume a training experiment endures $H$ hours on GPUs, with power consumption of $p_0$ kW, at the GPU utilization level (summed over all GPU nodes) $u(t)$ (in percent). Then the total energy consumption (in kWh) is given by  
\begin{eqnarray}
\begin{split}
\label{energyformula}
E = \int_0^H \frac{u(t)}{100} \cdot p_0 dt = H\cdot \bigg{(}\frac{1}{H}\int_0^H u(t) dt \bigg{)}\cdot \frac{p_0}{100}.
\end{split}
\end{eqnarray}
In practice, we sample $u(t)$ at the granularity of minutes throughout training using NVIDIA's System Management Interface (smi). We then calculate its cumulative sum $\overline{U}=\sum_{i=1}^{60H} u_i$, thereby we can approximate the right hand side of Equation \eqref{energyformula} by
\begin{eqnarray}
\begin{split}
\label{final_energy}
H\cdot\frac{\sum_{i=1}^{60H}u_i}{60H} \cdot \frac{p_0}{100}=\overline{U}\cdot \frac{p_0}{6000}. 
\end{split}
\end{eqnarray}
When reporting the energy consumption analysis for READ and other designs (see Section \ref{results_section}), we use $p_0=0.25$ kW for a NVIDIA V100 32 GB GPU \footnote{\href{https://www.nvidia.com/en-us/data-center/v100/}{250W comes from the datasheet on NVIDIA's website}} for Equation \eqref{final_energy}.

\subsection{Dataset and model details}
\label{appendix:data}
\textbf{GLUE Datasets} In Table \ref{data_size}, we list the dataset size, number of GPU nodes, and training batch size per GPU node for every task in GLUE. Note the total batch size (summed over all nodes) are fixed as $96$ across all tasks and all methods.

\textbf{\textbf{T5} models} Table \ref{model_size} gives architecture-related numbers for four sizes of $T_5$ model. Note for all experiments $T5_{\text{BASE}}$ we use the original archtectures, while for READ experiments with $T5_{\text{LARGE}}$, we drop the last $4$ layers from both encoder and decoder.

\begin{table}[]\centering
\caption{Model architectures for four different sized T5 models.}\label{model_size}
\begin{tabular}{
>{\columncolor[HTML]{FFFFFF}}c |
>{\columncolor[HTML]{FFFFFF}}c 
>{\columncolor[HTML]{FFFFFF}}c 
>{\columncolor[HTML]{FFFFFF}}c 
>{\columncolor[HTML]{FFFFFF}}c 
>{\columncolor[HTML]{FFFFFF}}c 
>{\columncolor[HTML]{FFFFFF}}c 
>{\columncolor[HTML]{FFFFFF}}c }
\toprule
                           & \begin{tabular}[c]{@{}c@{}}Params\\ (Million)\end{tabular} & \begin{tabular}[c]{@{}c@{}}Encoder \\  Layers\end{tabular} & \begin{tabular}[c]{@{}c@{}}Decoder \\  Layers\end{tabular} & Heads & \begin{tabular}[c]{@{}c@{}}Embedding\\ Dimension\end{tabular} & \begin{tabular}[c]{@{}c@{}}Head \\ Dimension\end{tabular} & \begin{tabular}[c]{@{}c@{}}FFN \\ Dimension\end{tabular} \\ \midrule
$\text{T5}_{\text{SMALL}}$ & 77                                                         & 6                                                          & 6                                                          & 8     & 512                                                           & 64                                                        & 2048                                                     \\
$\text{T5}_{\text{BASE}}$  & 248                                                        & 12                                                         & 12                                                         & 12    & 768                                                           & 64                                                        & 3072                                                     \\
$\text{T5}_{\text{LARGE}}$ & 771                                                        & 24                                                         & 24                                                         & 16    & 1024                                                          & 64                                                        & 4069                                                     \\
$\text{T5}_{\text{3B}}$    & 2885                                                       & 24                                                         & 24                                                         & 32    & 1024                                                          & 128                                                       & 16384                                                    \\ \bottomrule
\end{tabular}
\end{table}

\subsection{Hyperparameters}
\label{appendix:hp}

\begin{table}[]\centering\setlength\tabcolsep{2.9pt}
\caption{Final archtecture choices for all PEFT experiments reported in Section \ref{results_section}.}\label{arch-hyper}
\begin{tabular}{ccccccc}
\toprule
\rowcolor[HTML]{FFFFFF} 
{\color[HTML]{000000} }                                                                & {\color[HTML]{000000} READ}                                                                    & {\color[HTML]{000000} \begin{tabular}[c]{@{}c@{}}READ\\ large\end{tabular}}                    & {\color[HTML]{000000} Adapter}                                                  & {\color[HTML]{000000} LoRA}       & {\color[HTML]{000000} \begin{tabular}[c]{@{}c@{}}Prompt\\ tuning\end{tabular}} & {\color[HTML]{000000} LST}          \\ \midrule
\rowcolor[HTML]{FFFFFF} 
{\color[HTML]{000000} \begin{tabular}[c]{@{}c@{}}Architecture\\ HP Names\end{tabular}} & {\color[HTML]{000000} \begin{tabular}[c]{@{}c@{}}RNN-type/\\ RNN-dim\end{tabular}}             & {\color[HTML]{000000} \begin{tabular}[c]{@{}c@{}}RNN type/\\ RNN-dim\end{tabular}}             & {\color[HTML]{000000} \begin{tabular}[c]{@{}c@{}}Bottleneck\\ size\end{tabular}} & {\color[HTML]{000000} Rank}       & {\color[HTML]{000000} \begin{tabular}[c]{@{}c@{}}Number of\\ prompts\end{tabular}}   & {\color[HTML]{000000} Sidenet-dim}     \\ \midrule
\multicolumn{1}{l}{\begin{tabular}[c]{@{}l@{}}Architecture\\ Candidates\end{tabular}}  & \multicolumn{1}{l}{\begin{tabular}[c]{@{}l@{}}\{GRU/256,\\ GRU/128,\\ LSTM/128\}\end{tabular}} & \multicolumn{1}{l}{\begin{tabular}[c]{@{}l@{}}\{GRU/256,\\ GRU/128,\\ LSTM/128\}\end{tabular}} & \multicolumn{1}{l}{\{32, 64, 128\}}                                             & \multicolumn{1}{l}{\{8, 16, 32\}} & \multicolumn{1}{l}{\{10, 20, 30\}}                                             & \multicolumn{1}{l}{\{64, 96, 128\}} \\ \midrule
\rowcolor[HTML]{FFFFFF} 
{\color[HTML]{000000} \begin{tabular}[c]{@{}c@{}}Final\\ Choices\end{tabular}}         & {\color[HTML]{000000} GRU/256}                                                                 & {\color[HTML]{000000} GRU/256}                                                                 & {\color[HTML]{000000} 64}                                                       & {\color[HTML]{000000} 32}         & {\color[HTML]{000000} 20}                                                      & {\color[HTML]{000000} 96}           \\ \bottomrule
\end{tabular}
\end{table}

\begin{table}[]\centering\small
\setlength\tabcolsep{1.8pt}
\caption{Final learning rates for all fine-tuning methods and GLUE datasets}\label{lr-choices}
\begin{tabular}{
>{\columncolor[HTML]{FFFFFF}}c |
>{\columncolor[HTML]{FFFFFF}}c |
>{\columncolor[HTML]{FFFFFF}}c |
>{\columncolor[HTML]{FFFFFF}}c |
>{\columncolor[HTML]{FFFFFF}}c |
>{\columncolor[HTML]{FFFFFF}}c |
>{\columncolor[HTML]{FFFFFF}}c |
>{\columncolor[HTML]{FFFFFF}}c }
\toprule
{\color[HTML]{000000} }            & {\color[HTML]{000000} CoLA}                 & {\color[HTML]{000000} MNLI}                 & {\color[HTML]{000000} QNLI}                 & {\color[HTML]{000000} MRPC}                  & {\color[HTML]{000000} QQP}                  & {\color[HTML]{000000} SST-2}                & {\color[HTML]{000000} STS-B}                 \\ \midrule
{\color[HTML]{000000} Full-tuning} & {\color[HTML]{000000} $9\times 10^{-6}$}    & {\color[HTML]{000000} $7.16\times 10^{-5}$} & {\color[HTML]{000000} $3.76\times 10^{-4}$} & {\color[HTML]{000000} $3.59\times 10^{-5}$}  & {\color[HTML]{000000} $1.75\times 10^{-4}$} & {\color[HTML]{000000} $4.6\times 10^{-6}$}  & {\color[HTML]{000000} $1.30\times 10^{-4}$}  \\
{\color[HTML]{000000} Adapter}     & {\color[HTML]{000000} $1.16\times 10^{-3}$} & {\color[HTML]{000000} $7.47\times 10^{-4}$} & {\color[HTML]{000000} $4.6\times 10^{-6}$}  & {\color[HTML]{000000} $1.95\times 10^{-3}$} & {\color[HTML]{000000} $4.6\times 10^{-6}$}  & {\color[HTML]{000000} $1.46\times 10^{-4}$} & {\color[HTML]{000000} $2.83\times 10^{-3}$} \\
{\color[HTML]{000000} LoRA}        & {\color[HTML]{000000} $1.75\times 10^{-4}$} & {\color[HTML]{000000} $3.05\times 10^{-5}$} & {\color[HTML]{000000} $9\times 10^{-6}$}    & {\color[HTML]{000000} $1.75\times 10^{-4}$}  & {\color[HTML]{000000} $9\times 10^{-6}$}    & {\color[HTML]{000000} $7.16\times 10^{-5}$} & {\color[HTML]{000000} $1.15\times 10^{-4}$}  \\
BitFit                             & $3\times 10^{-3}$                           & $2.83\times 10^{-3}$                       & $2.83\times 10^{-3}$                       & $2.83\times 10^{-3}$                        & $2.83\times 10^{-3}$                       & $3\times 10^{-3}$                           & $2.83\times 10^{-3}$                        \\
Prompt-tuning                      & $2.83\times 10^{-3}$                       & $1.40\times 10^{-3}$                       & $7.47\times 10^{-4}$                        & $3\times 10^{-3}$                            & $2.83\times 10^{-3}$                       & $3\times 10^{-3}$                           & $2.74\times 10^{-3}$                        \\
LST                                & $2.51\times 10^{-4}$                        & $1.75\times 10^{-4}$                        & $7.16\times 10^{-5}$                        & $7.47\times 10^{-4}$                         & $3.7\times 10^{-4}$                         & $1.75\times 10^{-4}$                        & $1.45\times 10^{-3}$                        \\
READ                               & $3.29\times 10^{-4}$                        & $3.67\times 10^{-4}$                        & $1.75\times 10^{-4}$                        & $7.8\times 10^{-5}$                          & $1\times 10^{-6}$                           & $2.5\times 10^{-6}$                         & $4.6\times 10^{-5}$                          \\
READ-large                         & $8.5\times 10^{-5}$                         & $1.46\times 10^{-4}$                        & $1.75\times 10^{-4}$                        & $1.43\times 10^{-3}$                        & $2.13\times 10^{-4}$                        & $2.04\times 10^{-4}$                        & $7.1\times 10^{-5}$                          \\ \bottomrule
\end{tabular}
\end{table}

\textbf{Architecture search} For fine-tuning methods that have tunable architectural hyperparameters (e.g. RNN hidden dimensions in READ, ranks in LoRA, etc), we do hyperparameter search as follows: first, we fix the architecture $\mathcal{A}$ (e.g. in READ, take $\text{RNN-dim}=128$ and side-net type to be LSTM), and do learning rate search for every dataset $\mathcal{D}$. Among each hyperparameter sweep $\mathscr{H}(\mathcal{D})$ there exists a run $\mathcal{R}^*(\mathcal{D})$ that has the best validation score $\mathscr{S}(\mathcal{D})$. Then we calculate the average of $\mathscr{S}(\mathcal{D})$ across all datasets $\mathcal{D}$ as the quality score of $\mathcal{A}$, denoted as $\mathscr{S}(\mathcal{A})$. Now we move on to the next architecture (e.g. in READ, take $\text{RNN-dim}=256$ and side-net to be GRU) and repeat the above process. After iterating through all architecture candidates, we choose the architecture $\mathcal{A}^*$ that has the best score $\mathscr{S}(\mathcal{A}^*)$, and report the test scores of each best run $\mathcal{R}^*(\mathcal{D})$ of $\mathcal{A}^*$. Therefore, each method in Table \ref{tab:glue_results} adopts the same architectures throughout all datasets. For Full-tuning and BitFit where no architectural hyperparameters are present, we do the learning rate search once to obtain the test scores. 

\textbf{Learning rate search}
For each learning rate sweep, we do learning rates search in between $[1\times 10^{-6}, 3\times 10^{-3}]$ at log-scale for up to $32$ rounds, where we employ Bayesian optimization for faster convergence of hyperparameter sweeps at lower computation costs.


\textbf{Hyperparameter choices} Table \ref{arch-hyper} and \ref{lr-choices} summarize our final choices of architectural hyperparameters and learning rates.

\end{document}